\documentclass[conference]{IEEEtran}
\IEEEoverridecommandlockouts
\usepackage{cite}
\usepackage{amsmath,amssymb,amsfonts}
\usepackage{algorithmic}
\usepackage{graphicx}
\usepackage{textcomp}
\usepackage{xcolor}
\usepackage[hidelinks]{hyperref}

\newcommand{\myparagraph}[1]{\par\vspace{0.3em}\noindent\textbf{#1}\hspace{0.2em}}

\def\BibTeX{{\rm B\kern-.05em{\sc i\kern-.025em b}\kern-.08em
    T\kern-.1667em\lower.7ex\hbox{E}\kern-.125emX}}
\begin{document}


\title{Low-power event-based face detection with asynchronous neuromorphic hardware}

\author{
    \IEEEauthorblockN{
        Caterina Caccavella\IEEEauthorrefmark{1}\IEEEauthorrefmark{2},
        Federico Paredes-Vall\'es\IEEEauthorrefmark{1}
        Marco Cannici\IEEEauthorrefmark{2},
        Lyes Khacef\IEEEauthorrefmark{1}
    }
    \IEEEauthorblockA{
        \IEEEauthorrefmark{1}
        \itshape
        Stuttgart Laboratory 1, Sony Semiconductor Solutions Europe, Sony Europe B.V.
    }
    \IEEEauthorblockA{
        \IEEEauthorrefmark{2}
        \itshape
        Robotics and Perception Group, University of Zurich, Switzerland
    }
}

\maketitle


\begin{abstract}
The rise of mobility, IoT and wearables has shifted processing to the edge of the sensors, driven by the need to reduce latency, communication costs and overall energy consumption.
While deep learning models have achieved remarkable results in various domains, their deployment at the edge for real-time applications remains computationally expensive.
Neuromorphic computing emerges as a promising paradigm shift, characterized by co-localized memory and computing as well as event-driven asynchronous sensing and processing.
In this work, we demonstrate the possibility of solving the ubiquitous computer vision task of object detection at the edge with low-power requirements, using the event-based N-Caltech101 dataset. 
We present the first instance of an on-chip spiking neural network for event-based face detection deployed on the SynSense Speck neuromorphic chip, which comprises both an event-based sensor and a spike-based asynchronous processor implementing Integrate-and-Fire neurons. 
We show how to reduce precision discrepancies between off-chip clock-driven simulation used for training and on-chip event-driven inference. This involves using a multi-spike version of the Integrate-and-Fire neuron on simulation, where spikes carry values that are proportional to the extent the membrane potential exceeds the firing threshold.
We propose a robust strategy to train spiking neural networks with back-propagation through time using multi-spike activation and firing rate regularization and demonstrate how to decode output spikes into bounding boxes.
We show that the power consumption of the chip is directly proportional to the number of synaptic operations in the spiking neural network, and we explore the trade-off between power consumption and detection precision with different firing rate regularization, achieving an on-chip face detection mAP[0.5] of $\sim$0.6 while consuming only $\sim$20 mW.
\end{abstract}



\section{Introduction}
The growing interest in embedded edge devices and the increasing demand for more efficient solutions \cite{AIedgeRoadmap} requires the development of technologies that can operate at the edge, operating with low latency and low power \cite{Rabaey_etal19, Thompson_etal21}.
In this context, inspired by the remarkable capabilities of the human brain which stands as the most efficient processing system \cite{balasubramanian2021brain, sokoloff1960metabolism, siesjo1978brain}, neuromorphic computing emerges as a promising solution with a paradigm shift across sensor technology, hardware architectures, and algorithms. 

At the sensor level, event-based cameras are characterized by sparse and rapid data sampling, facilitating low latency in data transmission. In contrast to conventional frame-based and clock-driven sensors, event-based sensors discard redundant data, resulting in reduced power consumption since they only operate when changes in brightness are detected in the image space \cite{gallego2020event}. 
At the processor level, neuromorphic hardware architectures diverge from the traditional Von Neumann design, featuring co-located computing and memory components that operate asynchronously. 
On the algorithmic front, Spiking Neural Networks (SNNs), also referred to as the third generation of neural networks \cite{maass1997networks}, further enhance their suitability for sparse data containing temporal information. 

The compelling characteristics of neuromorphic computing have gained considerable interest, particularly within the field of computer vision. Extensive research has been dedicated to training SNNs on neuromorphic processors, resulting in a significant gain in the power consumption of the system for solving classification tasks \cite{ceolini2020hand,muller2022braille}. 
Conversely, off-chip training of SNNs has proven successful in tackling more complex regression tasks \cite{gehrig2020event,hagenaars2021self,cordone2022object}. However, most of these accomplishments remain constrained, solving simpler tasks in the former case and confined to simulation in the latter.
Accurate real-time inference is a stringent requirement for real-world vision tasks like object detection. Deploying networks on edge devices requires quick responsiveness and operation with a low-power footprint.

In this work we employ the second approach, namely off-chip training, and show that with a properly designed hardware-aware strategy that considers hardware constraints during training, deployment on dedicated neuromorphic hardware is possible for complex regression tasks.
We showcase our proposed design and training scheme for SNNs on the task of object detection at the edge using the SynSense digital neuromorphic processor Speck \cite{SpeckDataSheet}.
The hardware-aware algorithmic development of our SNN model is the main contribution of our work. It meticulously respects the various constraints imposed by the Speck chip, including considerations of network size, network activity and power consumption, addressing the ever increasing demand for low-power perception systems on the edge.


\section{Related Work}
\label{sec:related_work}  
\myparagraph{SNN training.} Developing SNNs for neuromorphic processors poses several challenges, with the primary one being the training of the SNN model. In this regard, training approaches can be categorized into conversion from Artificial Neural Network (ANN) to SNN and direct SNN training. The former strategy happens on simulation and involves simulating the SNN model on a conventional Von Neumann architecture, approximating the average firing rate of neurons with a ReLU activation \cite{eshraghian2021training},\cite{rueckauer2017conversion}. However, this approach has notable drawbacks, including a typical decline in model performance for spatio-temporal patterns and a high spike rate \cite{eshraghian2021training}.

The latter approach, i.e., direct training of SNNs, can be accomplished through local weight update methods, which are applicable on both simulation platforms \cite{serrano2013stdp} and neuromorphic hardware \cite{khacef2022spike}. Local plasticity approaches show promise in developing adaptive spiking models capable of learning from new data at the edge. Neuromorphic chips often lack support for global learning algorithms like backpropagation \cite{khacef2022spike}, making local plasticity crucial for on-chip learning.
Alternatively, SNN training can be achieved by employing standard back-propagation during simulation and subsequently deploying the trained model onto neuromorphic hardware for the inference phase, as described in \cite{cordone2022object}.
Surrogate gradient methods aim to address the inherent non-differentiability issue of spiking neurons. The method, as presented in \cite{neftci2019surrogate}, involves approximating the spiking function with a differentiable function in the backward pass, effectively enabling gradient-based optimization for SNNs.

SNNs trained directly with backpropagation methods can prove beneficial when offline training is sufficient, and the focus is on achieving low-latency and low-power inference at the edge. However, the training of deep SNN models presents challenges attributed to issues like vanishing and exploding gradients \cite{eshraghian2021training}. In standard neural network training, various strategies such as batch normalization, skip connections, dropout, and pooling are widely utilized to enhance stability and mitigate overfitting \cite{eshraghian2021training}. These strategies, designed for continuous-valued networks, often prove incompatible with the binary nature of SNNs and can pose challenges in terms of hardware compatibility with neuromorphic processors. Constraining the architecture of SNNs to be shallow and feed-forward has proven successful, with certain studies showcasing effective deployment of SNN models on neuromorphic sensors for classification tasks \cite{ceolini2020hand}, \cite{muller2022braille}. However, developing SNNs for computer vision tasks that are applicable to real-world scenarios necessitates the transition to more complex regression tasks.
Some successful works have focused on applications such as image reconstruction \cite{zhu2022event}, object tracking \cite{zhang2022spiking, xiang2022spiking}, optical flow estimation \cite{hagenaars2021self} and object detection \cite{kim2020spiking, cordone2022object, su2023deep}. However, these methods were investigated on simulation only and often rely on strategies that are not suitable for neuromorphic hardware. Notable exceptions to this involve hardware implementation on the Intel Loihi chip \cite{vitale2021event, paredes2023fully}. 

\myparagraph{SNN hardware and deployment.} Several types of neuromorphic processors have been developed, ranging from digital processors \cite{merolla2014million, davies2018loihi, orchard2021efficient}, mixed-signal and analog processors \cite{moradi2017scalable, indiveri2011neuromorphic}, and novel approaches based on memristive devices \cite{querlioz2013immunity, markovic2020physics}. All these processors have strict constraints, typically involving size limitations in the number of layers, neurons and connections, as well as processing constraints, including the specific methods used to implement operations such as pooling and skip connections.
Recent works \cite{cordone2022object} have shifted their focus towards directly training SNNs with the explicit goal of deploying the models on neuromorphic chips. This effort involves finding the sweet spot between incorporating successful strategies from ANN models and solutions that are compatible with neuromorphic hardware \cite{eshraghian2021training}.
The increasing focus on exploring the computational complexity of trained models has also gained interest \cite{yik2023neurobench}. This emphasis arises not only from the need to create chip-compatible models but also to develop them in a way that exploits the processing and communication advantages specific to the hardware.

Recent efforts have been made to propose variants of normalization strategies that can be used during the training phase and eliminated or integrated at inference, after the model is deployed on a neuromorphic chip \cite{zheng2021going, cordone2021learning}. Moreover, many works pay close attention to quantifying the number of operations and parameters required by the model \cite{zhu2022event, cordone2022object, su2023deep}, which directly correlates with power consumption \cite{Abderrrahmane_etal20}. For instance, in \cite{kim2020spiking}, a scaled-down YOLOv1 with batch normalization and max-pooling is used for object detection, with energy efficiency evaluated on a GPU. In \cite{su2023deep}, a more advanced YOLO version employs fully spiking residual blocks. In \cite{cordone2022object}, the SSD framework is used, successfully implementing skip connections with binary operations. They also propose smaller architectures and introduce a normalization mechanism in SNN for future on-chip inference. Despite the comprehensive development of SNN architectures in these studies, tailored for compatibility with neuromorphic processors and providing hardware cost metrics, the actual deployment of the model on neuromorphic processors often remains unexplored. As a result, the solutions proposed by these works remain largely on simulation and lack concrete implementation on neuromorphic hardware.


\section{Preliminaries}
\subsection{Event-based cameras}
\label{sub:evs}
Event-based cameras react to changes in brightness in the scene by providing as output a spatio-temporal stream of sparse and independent events. Each event is defined as $\mathbf{e}_i = (x_i, y_i, t_i, p_i)$, where $x_i,y_i$ are the coordinates of the pixel, $t_i$ is the timestamp at microseconds resolution and $p_i\in \{-1, 1\}$ is the polarity of the event, i.e., whether the brightness increased or decreased.
The unique data streaming method employed by these sensors, as opposed to traditional 2D camera frames, necessitates data adaptation to meet the requirements of standard deep learning frameworks and computer vision techniques when operating on conventional processors.
To address this challenge, the simplest approach is to aggregate incoming events into frames, converting event-based data into a frame-based format compatible with deep learning frameworks \cite{gallego2020event}. Achieving the optimal representation requires striking the balance between the number of frames used, impacting power and memory usage, and effectively harnessing the temporal information from the event-based stream to enhance performance.
Some of the most common aggregation strategies used in event-based data processing include the histogram of events \cite{maqueda2018event}, which counts the number of events within a fixed time interval, and the binary image representation which simply stores a 1 if there is any event detected within that interval and a 0 otherwise.
The challenge of integrating event-based sensors with traditional processing tools extends beyond event representations. Asynchronous event data is best suited for asynchronous processors \cite{davies2018loihi}, while standard Von-Neumann architectures with clocked circuits can lead to redundant computation and processing limitations. A promising solution is to combine event-based sensors with asynchronous processors that share the same computational principles.

\subsection{SynSense Speck neuromorphic processor}
\label{subsec:speck_architecture}

\begin{figure}[t]
\centerline{\includegraphics[width=\linewidth]{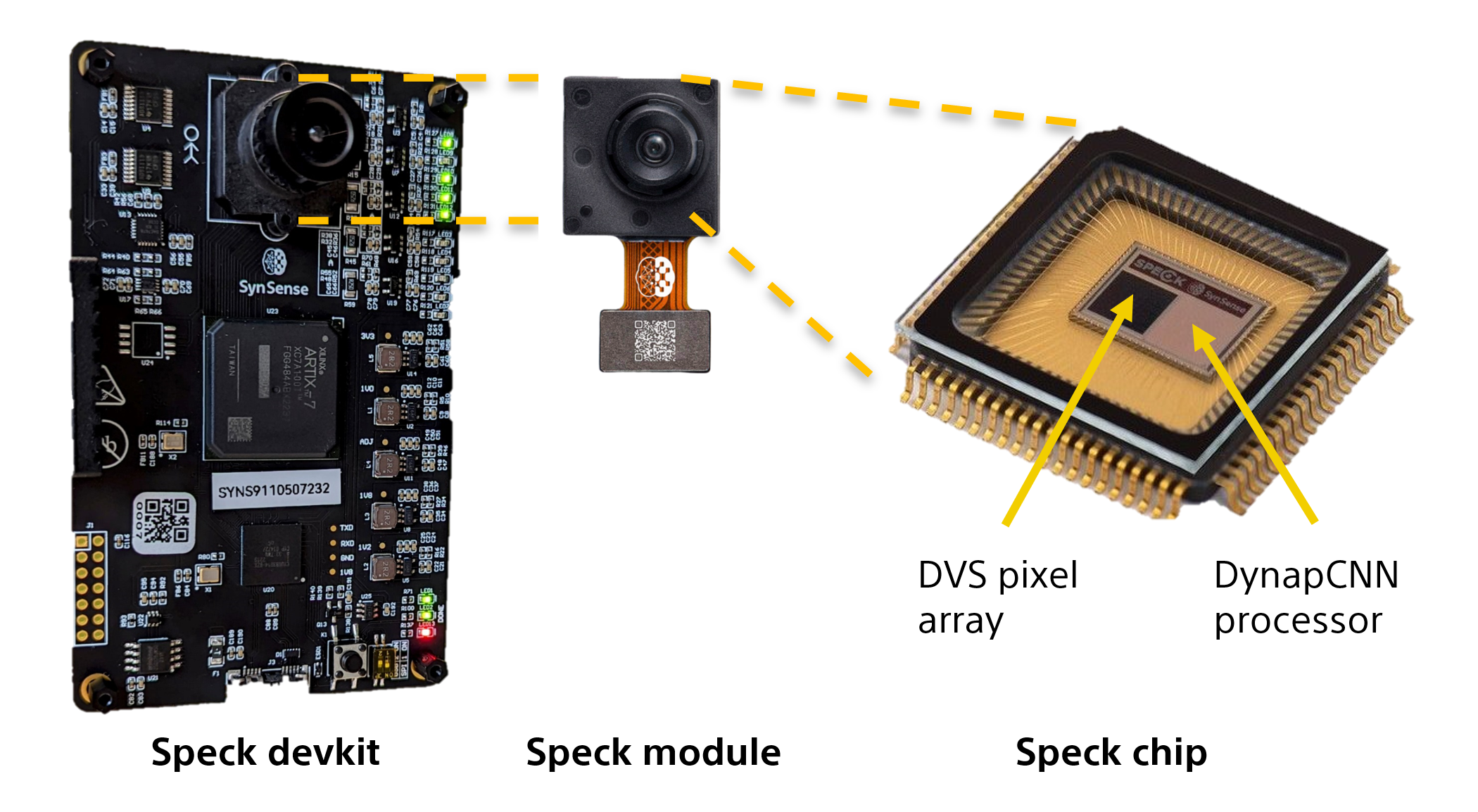}}
\caption{Speck Development Kit (left) integrates the Dynamic Vision Sensor (DVS) (middle) with the asynchronous neuromorphic processor DynapCNN (right).}
\label{fig:dev_kit}
\end{figure}

The processor employed in this study is the digital neuromorphic processor Speck \cite{SpeckDataSheet}, developed by SynSense\footnote{https://www.synsense.ai/}. The Speck chip integrates an event-based vision sensor and an asynchronous neuromorphic processor, as shown in Fig.\ \ref{fig:dev_kit}, with the primary objective of processing events using convolutional SNNs. 
The Speck chip can receive input events either from the embedded sensor featuring a pixel array with a resolution of $128 \times 128$ or from data stored off-chip, sent as a sequence of $(x, y, t, p)$ events, see Section \ref{sub:evs}. Incoming inputs are queued and sent to the processing cores in a First-In-First-Out (FIFO) fashion. The processing cores are limited to nine convolutional cores, each executing a per-event computation sequence of convolution $\rightarrow$ Integrate-and-Fire (IF) spiking neuron $\rightarrow$ sum pooling. These cores operate asynchronously and independently, without any synchronization mechanism or global timesteps. The FPGA in the development kit is responsible for timestamping outgoing events, facilitating conversion of the output into a time-based format if necessary. 

Each neural core in the chip has distinct memory constraints, categorized into synaptic weights memory (8-bit precision) and neurons states memory (16-bit precision). Additionally, each core operates under a limited bandwidth constraint, quantified in terms of the number of synaptic operations per second (SynOps/s), which is calculated by the number of spikes produced by a neuron multiplied by its fanout. When the maximum SynOps/s capacity is exceeded, events may be delayed or dropped, therefore maintaining low firing activity among the spiking neurons is crucial. 
The output events from the Speck chip can be directly retrieved from the convolutional cores at a frequency of 25 MHz or from the dedicated readout cores. However, the latter is limited to only 16 neurons, making it unsuitable for our application.

The challenge of deploying the SNN on Speck chip derives from (1) the size and memory constraints of the convolutional cores, (2) the quantization of the network parameters, (3) the maximum SynOps/s per layer and (4) the difference in the per event-based processing mechanism of Speck compared to the timestep-based processing of conventional CPU/GPU.
The network is trained off-chip, and then deployed on-chip after weight quantization. This comes at the cost of a sim-to-real mismatch, as the stream of input events needs to be quantized at specific $\delta t$ intervals on simulation.
While this gap could be alleviated by simulating the per-event processing behavior of the chip, this is unfeasible in practice as it would require simulating a very high number of timesteps, resulting in unrealistic training times for BackPropagation Through Time (BPTT) with high GPU memory usage \cite{taylor2023addressing}.


\section{Methods}
Efficiently deploying a model on the chip is tied to the chip's design constraints. The challenges fall into two main categories: model topology and training strategy. 
The first challenge stems from the chip's core architecture, supporting only IF neurons and imposing strict memory limits. Addressing this involves restricting the network topology to a maximum of 9 convolutional layers, and precisely adhering to the kernel and neuron memories constraints. 
The second challenge arises from the differences in processing mechanisms between CPU/GPU and the Speck processor, leading to issues related to the sim-to-real gap. Moreover, models can easily saturate the available chip cores bandwidth of SynOps/s. In Section \ref{subsec:neuron_model}, we mitigate the sim-to-real shift by adopting the approach of \cite{weidel2021wavesense} to approximate on-chip behavior, while, in Sections \ref{subsec:normalization} and \ref{sec:activity_reg}, we propose a regularization training strategy to limit on-chip bandwidth, enabling the successful deployment of detection networks on-chip.

\subsection{Training of multi-spike Integrate-and-Fire model}
\label{subsec:neuron_model}
SNNs take close inspiration from the behavior and mechanisms of biological neurons, relying on sparse and binary spikes for information processing. In this study, we use the simple IF spiking neuron, constrained by the Speck processor. 
For each incoming spike, the corresponding synaptic weight is accumulated in the membrane potential of the neuron $U(t)$. If the membrane potential $U(t)$ exceeds the firing threshold, the neuron fires an output spike and the membrane potential is reset by subtracting the membrane potential (i.e., soft reset). The general formulation is defined by Eqs.\ \ref{eq:LIFneuron_dl} and \ref{eq:LIFneuron_dl_spk}:

\begin{equation}
\label{eq:LIFneuron_dl}
    U(t) = U(t-1)+WX(t)-S_{out}(t-1)\theta,
\end{equation}
\begin{equation} \label{eq:LIFneuron_dl_spk}
S_{out}(t) =
\left\{
\begin{aligned}
    & 1, \quad \text{if } U(t) > \theta \\
    & 0, \quad \text{otherwise}
\end{aligned}
\right.
\end{equation}

Here, $U(t)$ is the neuron's membrane potential, $W$ is the input connection weights, $X(t)$ is the neuron input, $S_{out}(t)$ are the generated spikes, and $\theta$ is the membrane potential threshold.
The spike generation mechanism described above involves the generation of a single spike when the membrane potential $U(t)$ of a neuron exceeds the membrane potential threshold $\theta$. While this mechanism represents the formal definition of the IF neuron model, it may be necessary to consider variants of this mechanism in certain situations such as deploying a trained network onto a neuromorphic chip with per-event processing, as presented in \cite{weidel2021wavesense}. 

As discussed in Section \ref{subsec:speck_architecture}, the divergent behavior of the SNN in simulation and on the Speck chip can significantly affect the model's precision. The need to discretize the stream of input events in temporal bins during simulation leads to a reduction in the number of spikes produced by the neurons compared to the on-chip inference where each event is processed independently \cite{weidel2021wavesense}.
The single-spike model might undergo two main problems when deployed on the Speck chip. First, if the time interval for event aggregation used in simulation is not sufficiently small, the on-chip inference might generate a significantly higher number of spikes for the same input. This may lead the chip to stall due to limited core bandwidth. Second, even if the number of SynOps/s does not reach levels that cause chip stalls, the on-chip precision would experience a substantial drop. We propose to address this issue following the multi-spike approach in \cite{weidel2021wavesense}.

In the multi-spike generation mechanism, the number of spikes generated when the membrane potential exceeds the threshold becomes directly proportional to the membrane potential value. This adjustment allows for the use of a large enough timestep during the training, thus enabling a reasonable simulation time. The mathematical representation of this modification is defined by Eq.\ \ref{eq:multi_spike_eq}:

\begin{equation}
\label{eq:multi_spike_eq}
S_{out}(t) =
\left\{
\begin{aligned}
    & \left\lfloor \frac{U(t)}{\theta} \right\rfloor, \quad \text{if } U(t) > \theta \\
    & 0, \quad \text{otherwise}
\end{aligned}
\right.
\end{equation}

The choice of the spiking generation function significantly influences the training process. In our experiments, we use the surrogate gradient method \cite{neftci2019surrogate} for training, which approximates the step activation function with a differentiable approximation during the backward pass.
This substitution enables gradient computation and facilitates weight updates throughout the network. In the case of multi-spike activation, following \cite{weidel2021wavesense}, we substitute the standard surrogate function with a periodic form, which accommodates the multi-spike behavior.
In summary, we use a single exponential with the single-spike generation function and a periodic exponential with the multi-spike generation function. The surrogate gradient is employed for weight updates through BPTT \cite{werbos1990backpropagation}.

\subsection{Spike values normalization}
\label{subsec:normalization}
In order to tackle regression tasks such as object detection, the output spike events need to be converted into continuous values. To accomplish this conversion, we have chosen to employ a linear (i.e., fully connected) layer implemented off-chip \cite{hagenaars2021self, paredes2023fully}.
However, given the unbounded nature of the multi-spike generation function, the model may exhibit significantly high values at the output. This, coupled with the sparse activation of SNN layers, could lead to high internal variance at the output of the on-chip spiking model, causing unstable gradient flow.

To handle the on-chip output spikes effectively, we added a layer normalization step off-chip. Layer normalization \cite{ba2016layer}, introduced to overcome batch normalization limitations, normalizes output across features, eliminating batch dependence. We demonstrate the effectiveness of this technique in the context of SNN deployment. The inherent statefulness of SNNs typically results in a larger memory footprint during training due to BPTT acting not only over the spatial dimensions but also over time. This often leads to the use of small batch sizes, posing a challenge when normalization depends on batch statistics. Additionally, during the deployment of SNNs for the inference phase, batches may not be used, making the normalization over features a viable alternative that eliminates the need for different computations in the training and inference phases as observed with batch normalization.


\label{eq:layer_normalization_combined}

\subsection{Internal activity regularization}
\label{sec:activity_reg}
The training of deep SNNs often introduces challenges related to vanishing or exploding gradients \cite{eshraghian2021training}. In our specific context, precise control over firing rates holds particular significance for the accurate deployment of the network on the chip, as elaborated in Section \ref{subsec:speck_architecture}. Inspired by \cite{zenke2021remarkable, rossbroich2022fluctuation}, we propose to solve this issue by integrating the hardware bandwidth constraints during training, in the form of firing rate regularization.

We regulate the firing rates of neurons by adding a penalization term to the detection loss function with the aim of preventing excessive firing activity. Specifically, we choose L1 regularization which penalizes the number of spikes in each layer and is defined in Eq.\ \ref{eq:ref_loss}:

\begin{equation}
\label{eq:ref_loss}
L_1 = \sum_{t}^T \lambda \left(\sum_{l}^L \left( \sum_{i}^{N_l} F_{l,i}[t] \right)\right).
\end{equation}

Here, $T$ is the number of timesteps of a sample, $L$ is the number of layers, $N_l$ is the number of neurons in each layer $l$, $\lambda$ controls the importance of the penalization loss, and $F_{l,i}$ denotes the firing rate of layer $l$ and neuron $i$.


\section{Implementation details}

\subsection{Model topology}
\label{sub:model_topology}

\begin{figure}
\centerline{\includegraphics[width=\linewidth]{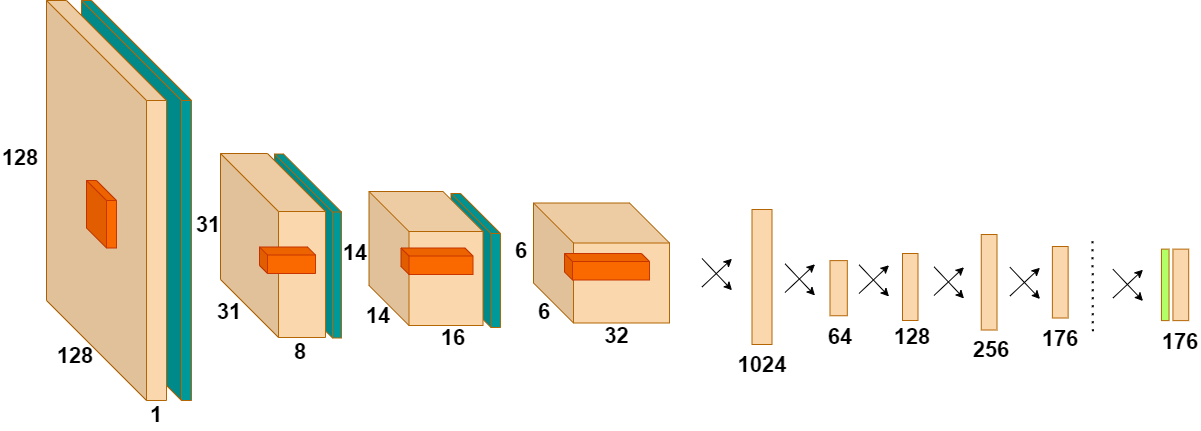}}
\caption{Topology of the Speck-compatible model. The layers to the left of the dotted line represent the on-chip spiking model, while those to the right represent off-chip layers. All convolutional kernels have a size of 3x3.}
\label{fig:model}
\end{figure}

The deployment of the SNN on Speck introduces important limitations in terms of network design. In this work, we present two YOLO-inspired architectures \cite{redmon2016you}. 

The first model is our adaptation of the YOLE model \cite{cannici2019asynchronous}, where we replaced the LeakyReLU activation function with a simple ReLU activation. This modification is motivated by the similarity of the multi-spike activation function used in our SNN model to a quantized version of the ReLU function. This architecture is referred to as ANN-YOLE when utilizing the ReLU activation function and SNN-YOLE-SS with a single-spike activation function. 
The second model, shown in Fig.\ \ref{fig:model} is specifically designed to fulfill the Speck chip constraints, consisting of 4 convolutional and 4 fully-connected layers without biases. Channels and kernel dimensions are determined to comply with kernel and neuron memory constraints while maximizing input resolution at $128 \times 128$. The model utilizes $66\%$ of the total kernel memory and $18\%$ of the total neuron memory on the chip, reaching the maximum memory available for some of the cores. This architecture is referred to as ANN-Speck when utilizing the ReLU activation function, SNN-Speck-SS with a single-spike activation function, and SNN-Speck-MS with a multi-spike activation function.

All the layers in the model are composed of IF neurons which can be simulated with either a single-spike or a multi-spike activation, as described in Section \ref{subsec:neuron_model}. Although the Speck-YOLO architecture can be effectively deployed on-chip, the output from the last layer is in the form of event spikes. 
In order to tackle the regression task, we employ a linear layer with biases implemented off-chip on the CPU, preceded by layer normalization for the multi-spike models.

\subsection{Dataset}
\label{subsec:Ncaltech101}
The N-Caltech101 dataset \cite{orchard2015converting} was chosen to conduct the experiments. This dataset comprises event recordings of images representing 100 different classes, and an additional background class which is excluded in our experiments. The dataset is created by recording static images using an ATIS event-based camera \cite{posch2010qvga}, performing three saccadic movements. Each sample in the dataset contains $(x,y,t,p)$ events and four ground truth bounding boxes, originally provided in coordinates relative to the sensor resolution. To adapt them for our use case, we shifted and converted the ground truths to the PASCAL VOC format \cite{everingham2010pascal}, resulting in the ground truth \((x_{\text{min}}, y_{\text{min}}, x_{\text{max}}, y_{\text{max}}, \text{objectness}, \text{label})\). Here, $x_{\text{min}}$, $y_{\text{min}}$, $x_{\text{max}}$, $y_{\text{max}}$ denote the top-left and bottom-right corners of the bounding box, ``objectness" represents the confidence score (set to 1), and ``label" is the class index (0 to 99).

The N-Caltech101 dataset presents significant challenges due to the high number of classes and uneven sample distribution among classes. This makes dataset splitting a critical aspect of the problem \cite{cannici2019asynchronous}. We divided the dataset into training and validation sets using a stratified split of 80\% for training and 20\% for validation. Furthermore, we augment the dataset by applying random scaling and cropping to the input frames, following the approach in \cite{cannici2019asynchronous}. 
The entire dataset is utilized at a resolution of $240 \times 240$ for the experiments involving the YOLE-based model. The multi-class problem causes the last layer of the model to exceed the constraints from the chip. Thus, we move to a single-class problem where only the ``faces'' class is selected. The events in each sequence are down-sampled to a resolution of $128 \times 128$ to meet the input resolution requirements of the on-chip implementation.

To train the model, unless specified otherwise, we use the binary event representation in combination with the single-spike activation and the histogram of events in combination with the multi-spike activation. Note that the former only provides information about the per-pixel presence/absence of events in a given time window, while the latter counts the events in this interval. Hence, the histogram of events is expected to reduce the sim-to-real gap between our timestep-driven simulation and the per-event processing done on-chip. 


\section{Experiments}

\subsection{YOLE baseline on N-Caletch101}
\label{subsec:yole_ncaltech101}
In this section, we present the object detection results of the adapted YOLE models on N-Caltech101. The input samples are divided into window lengths of 90 ms, with each bin corresponding to one saccade. The image representation used is the histogram of events for the ANN and binary images of events for SNN with single-spike activation. The models are trained for 100 epochs with a batch size of 40. Results are shown in Tab.\ \ref{table:ncaltech101_benchmark}.

\begin{table}
\caption{Detection precision on simulation of the baseline YOLE and our models on the N-Caltech101 dataset.}
\begin{center}
    \begin{tabular}{lccr}
        \hline
        Model & Activation & mAP[0.5] \\
        \hline
        YOLE \cite{cannici2019asynchronous} & LeakyReLU & 0.398 \\
        ANN-YOLE (Ours) & ReLU & 0.346  \\
        SNN-YOLE-SS (Ours) & SS IF neuron & 0.259 \\
        \hline
    \end{tabular}
\label{table:ncaltech101_benchmark}
\end{center}
\end{table}

The ANN-YOLE model with default parameters achieves an mAP[0.5] of 0.346, while the baseline model from \cite{cannici2019asynchronous} achieves a performance of 0.398. This discrepancy can be attributed to the fact that the YOLE baseline model employs a pre-training stage, where the backbone is trained on classifying N-Caltech101 objects, and fine-tunes loss-related parameters to achieve a better performance. On the contrary, we do not perform pre-training and use the hyper-parameters from \cite{redmon2016you} directly.
The SNN model achieves an mAP[0.5] of 0.259 in the single-spike case. In comparison to the ANN version, the SNN demonstrates a marginally reduced detection loss, but a higher classification loss which reduces the overall mAP.

The mAPs[0.5:0.95] (not shown in Tab.\ \ref{table:ncaltech101_benchmark}) are very similar, being 0.148 for ANN-YOLE and 0.142 for SNN-YOLE-SS, which suggests that the SNN performs better with some of the higher IoU thresholds. 
Further training to reach such precision is not pursued due to the N-Caltech101 dataset's inherent challenges when considering deployment on the Speck processor, such as the high number of classes that would demand an architecture exceeding chip constraints. Therefore, we limit our scope to the single-class problem of face detection.

\subsection{Speck-YOLO for face detection}
\label{subsec:speck_yolo}
In this section, we report the performance of the YOLE model on simulation for the class ``faces'' from N-Caltech101. Additionally, we address the same task using the Speck-YOLO model described in Section \ref{subsec:speck_architecture}. The results are presented in Tab.\ \ref{table:faces_benchmark}.
We use a window length of 90 ms, for comparison with previous results from literature, and 10 ms for the new experiments to enhance the simulation of real-time scenarios with rapid changes in the data stream. The models are trained for 100 epochs with a batch size of 16.

The ANN-Speck model exhibits very similar performance compared to the bigger ANN-YOLE model, indicating that the smaller model is sufficient for solving the single-object detection task. Furthermore, we observe that for a larger time window of 90 ms, the ANN models outperform the SNN model. This may be attributed to the input representation employed, which contains limited temporal information.
However, one might expect this to hold true for a time window of 90 ms but not for 10 ms, which contains fewer events and could be insufficient for solving the task with a stateless ANN. As expected, we observe that the ANN's performance decreased with the smaller 10 ms window dropping from an mAP[0.5] of 0.883 to an mAP[0.5] of 0.775 although it still outperforms the single-spike SNN which reached only an mAP[0.5] of 0.587.
However, note that the SNN with multi-spike activation functions, characterized by an mAP[0.5] of 0.923, outperforms both the single-spike SNN and the baseline ANN for the 10 ms window. We hypothesize that this is mainly due to the high activation range of the multi-spike activation and its stateful processing of temporal information.

\begin{table}
\caption{Face detection precision on simulation using the ``faces'' class from the N-Caltech101 dataset.}
\begin{center}
    \begin{tabular}{lccrc}
        \hline
        Model & Activation & Window length & mAP[0.5]\\
        \hline
        ANN-YOLE  & ReLU & 90ms & 0.871 \\
        ANN-Speck & ReLU & 90ms & 0.883 \\
        ANN-Speck & ReLU & 10ms & 0.775 \\
        SNN-Speck-SS & SS IF neuron & 90ms & 0.715\\
        SNN-Speck-SS & SS IF neuron & 10ms & 0.587 \\
        SNN-Speck-MS & MS IF neuron (LN) & 10ms & 0.923 \\
        \hline
    \end{tabular}
\label{table:faces_benchmark}
\end{center}
\end{table}

\begin{table}
\caption{
Impact of different normalization strategies on the face detection precision on simulation of the SNN-Speck model with multi-spike activation using the ``faces'' class from the N-Caltech101 dataset.}
\begin{center}
    \begin{tabular}{lccrc}
        \hline
        Model  & Time window & Normalization & mAP[0.5] &\\
        \hline
        SNN-YOLE-MS  & 90ms & - & 0.233  \\
        SNN-YOLE-MS  & 90ms & Batch norm & 0.05   \\
        SNN-YOLE-MS  & 90ms & Layer norm & 0.969   \\
        SNN-Speck-MS & 10ms & - & 0    \\
        SNN-Speck-MS & 10ms & Batch norm & 0.001  \\
        SNN-Speck-MS  & 10ms & Layer norm & 0.923  \\
        \hline
    \end{tabular}
\label{table:LN}
\end{center}
\end{table}

\begin{table*}
    \caption{
    Impact of firing rate regularization on the face detection precision and power consumption of Speck-compatible models on the ``faces'' class from the N-Caltech101 dataset. Note that precision is measured both on simulation (using floating-point and quantized versions of the models) and on chip (also quantized). A dash indicates a value not retrieved due to chip stalls.}
    \begin{center}
    \begin{tabular}{lcccccccc}
        \hline
        \begin{tabular}[c]{@{}c@{}}Spiking\\ activation\end{tabular} & Reg. & \begin{tabular}[c]{@{}c@{}}Sim. \\ mAP[0.5]\end{tabular} & \begin{tabular}[c]{@{}c@{}}Sim. Quant. \\ mAP[0.5]\end{tabular} & \begin{tabular}[c]{@{}c@{}}Chip \\ mAP[0.5]\end{tabular} & \begin{tabular}[c]{@{}c@{}}\begin{tabular}[c]{@{}c@{}}Chip \\ Power (mW)\end{tabular}\end{tabular} & \begin{tabular}[c]{@{}c@{}}Chip\\Spikes/s\end{tabular} & \begin{tabular}[c]{@{}c@{}}Sim.\\Spikes/s\end{tabular} & \begin{tabular}[c]{@{}c@{}}Sim.\\SynOps/s(M)\end{tabular}\\
        \hline
        Speck-SS & 0 & 0.838 & 0.845 & - & - & - & - & - \\
        Speck-SS & 5e-3 & 0.620 & 0.640 & 0 & - & - & - & - \\
        Speck-MS & 0 & 0.923 & 0.919 & - & - & - & 26,390,291 & 464 \\
        Speck-MS & 1e-4 & 0.864 & 0.870 & 0.868 & 33.2 & 567,002 & 1,982,899 & 39.1 \\
        Speck-MS & 1e-3 & 0.763 & 0.763 & 0.780 & 33.1 & 553,794 & 472,773 & 9.6 \\
        Speck-MS & 2e-3 & 0.685 & 0.717 & 0.705 & 24.8 & 411,300 & 307,160 & 6.3 \\
        Speck-MS & 3e-3 & 0.628 & 0.623 & 0.622 & 19.4 & 319,966 & 231,079 & 4.6 \\
        Speck-MS & 4e-3 & 0.566 & 0.577 & 0.598 & 17.3 & 279,745 & 199,202 & 4.0 \\
        Speck-MS & 5e-3 & 0.527 & 0.546 & 0.565 & 11.8 & 192,292 & 146,441 & 2.9 \\
        Speck-MS & 1e-2 & 0.407 & 0.406 & 0.419 & 7.3 & 104,767 & 78,299 & 1.7 \\
        \hline
     \end{tabular}
     \label{table:ms_regularization}
     \end{center}
\end{table*}

\begin{figure*}
    \centering
    \includegraphics[width=\textwidth]{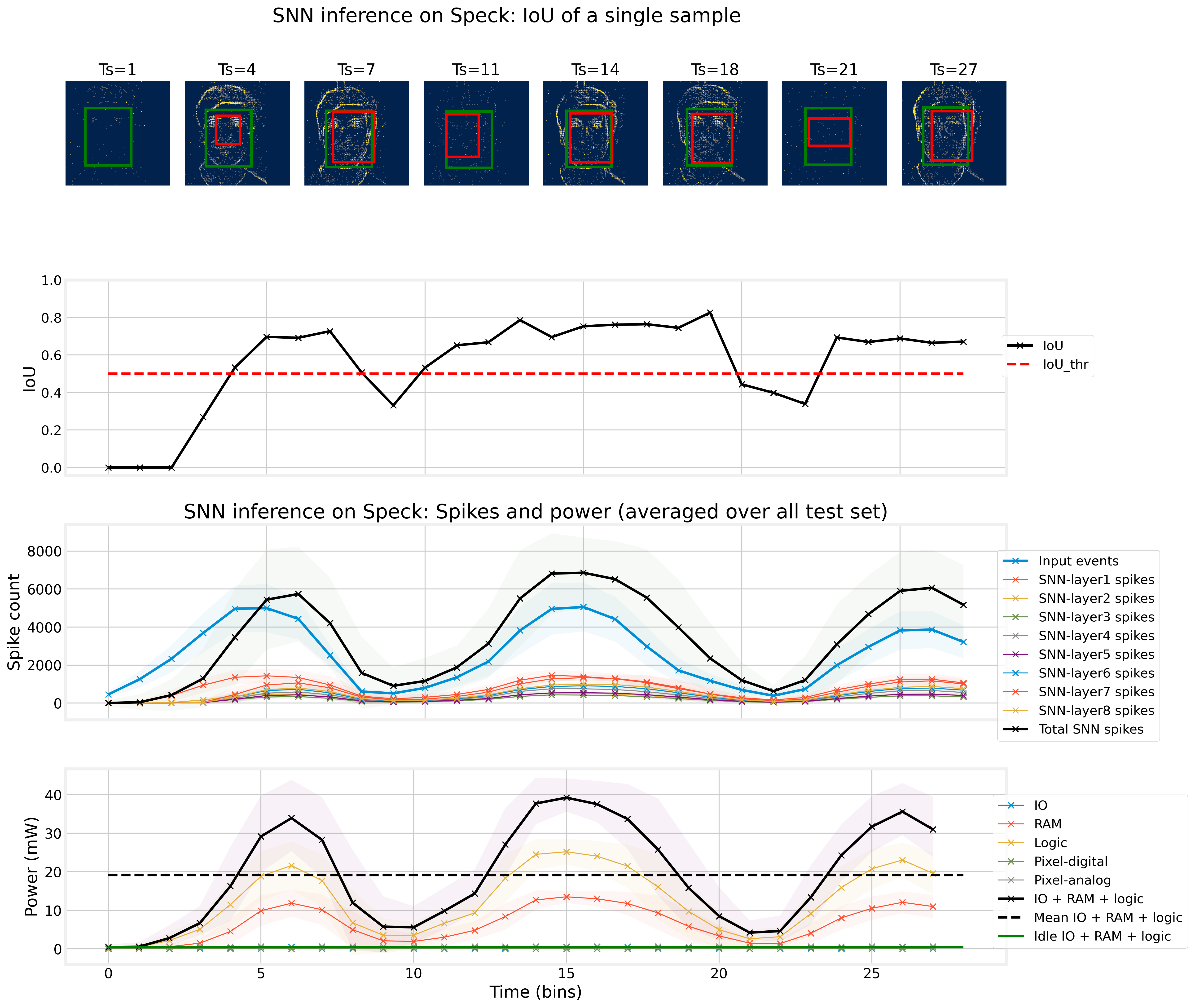}
    \caption{On-chip inference of SNN-Speck model with multi-spike activation. On top, the IoU between the correct and predicted bounding boxes for each timestep within a single sample. In the middle, the number of input events and on-chip spikes of the best model. On the bottom, the power consumption of the on-chip inference of the best model.}
    \label{fig:IoU}
\end{figure*}

\subsection{The problem with deploying single-spike SNNs on Speck} 
\label{subsec:ss_vs_ms}
The model trained with single-spike activation is expected to have a poor performance when deployed on the chip, as discussed in Section \ref{subsec:speck_architecture}. 
To investigate this behavior further, we conducted a comparative analysis by testing in simulation the single-spike model with an equivalent model using multi-spike activations.
All the models are trained with a window length of 10 ms during 100 epochs with a batch size of 16. 

Even though the neuron models in the Speck follow the standard (i.e., binary) IF model, the mismatch between simulation and chip in terms of the processing of the input events causes networks trained with single-spike activations to be not suitable for chip deployment (see Section \ref{subsec:neuron_model}).
To test this, we train a model with single-spike activation and binary input representation. Subsequently, we test in simulation the same model by substituting the single-spike with a multi-spike activation and the input representation with a histogram of events. 
This test aims to observe the implications of deploying a single-spike model on-chip by approximating the per-event processing on simulation.

As expected, testing the single-spike model with a multi-spike activation results in a substantial increase in the number of spikes, transitioning from a maximum of 8000 spikes to an elevated count of 14 million. Note that this is far beyond the maximum firing rate supported by the Speck chip and would cause chip stalls if the model was to be deployed.
These results are confirmed and further analyzed in Section \ref{subsec:regularization}.



\subsection{Spikes value normalization for multi-spike SNNs} 
\label{subsec:ms}
Training the model with the multi-spike activation function mitigates the sim-to-real gap, yet it introduces a challenge related to high firing rates. The resulting spike tensors usually display elevated values and high internal variance. To address this, we normalize the values of the last layer before inputting them into the decoding network, as discussed in Section \ref{subsec:normalization}.
Batch normalization can partly mitigate the problem, but it introduces oscillations during model convergence and is sensitive to variations in initialization and training parameters. On the other hand, the proposed way of normalizing values over features instead of batches provides greater stability and robustness. Layer normalization proves to be an effective technique.

Table \ref{table:LN} presents a summary of results for both single-spike and multi-spike models trained under various conditions, including scenarios without normalization, with batch normalization and with layer normalization. 
Notably, layer normalization demonstrates robustness across different architectures. Its effectiveness is evident in the significant improvement of the SNN-YOLE architecture, increasing from an mAP[0.5] of 0.233 without normalization to 0.969 with layer normalization. Similarly, the SNN-Speck model progresses from an mAP[0.5] of 0 without normalization to 0.923 with layer normalization, highlighting the importance of this normalization strategy.



\subsection{SNN firing rate regularization for chip deployment}
\label{subsec:regularization}



\begin{figure}[t]
\centerline{\includegraphics[width=0.49\textwidth]{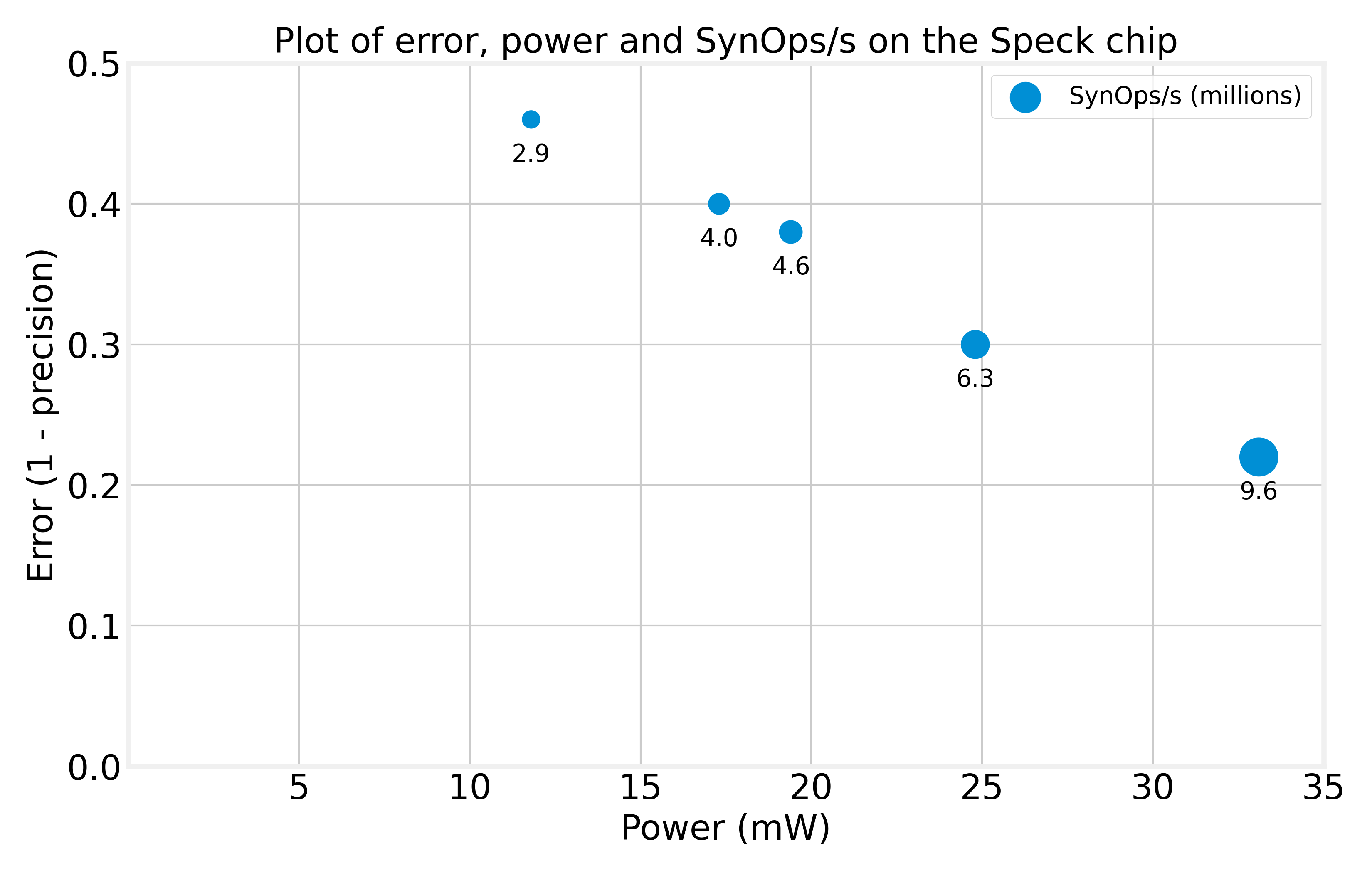}}
\caption{Pareto-like curve between error and power consumption. The area of the circles represents the averaged number of SynOps/s. Bottom-left region is the best tradeoff.
}
\label{fig:pareto_synops_plot}
\end{figure}

In this section, we present the results of the SNN-Speck model for the face detection task on-chip, which was trained with varying levels of regularization on the firing rate.
In addition to using mAP as a performance metric, it is crucial to consider the firing rate of the model for two primary reasons. Firstly, the number of spikes directly impacts the model's power consumption, making it essential to keep the firing rate low. Secondly, each core of the Speck has a maximum bandwidth. If this limit is exceeded, events may be dropped or delayed (see Section \ref{subsec:speck_architecture}).
To address the issue of high firing rates and reduce the number of synaptic operations, we introduced an L1 penalization term on the neurons' firing rates during training, as discussed in Section \ref{sec:activity_reg}. We experimented with different penalization parameters. The resulting mAP scores are presented in Tab.\ \ref{table:ms_regularization}. Notably, as the strength of the penalization increases, the number of spikes decreases, but this reduction in firing activity also decreases precision.

Table \ref{table:ms_regularization} provides insights into mAP scores before model quantization, after model quantization, and on the chip. Additionally, we present the running average power consumption of the model probed at 100 Hz, the average number of spikes generated by the model on chip per second, the average number of spikes generated by the model in simulation per second, and the average number of SynOps/s generated by the model in simulation.

First, we present the comparison between the SNN-Speck-SS trained without any firing rate penalization term and the model trained with a firing rate penalization term of $5e-3$, intended to mitigate excessive internal activity. The firing rate of the model trained without regularization tends to reach excessively high values, leading to chip stalls and rendering event retrieval impossible. Conversely, penalizing high internal activity helps prevent chip stalls, but it comes at the cost of a notable decline in model precision. As thoroughly explained, the adoption of multi-spike training is crucial to address this discrepancy.

Without any regularization strategy, the multi-spike  model presents a very high firing rate which causes the maximum bandwidth of the cores to be reached and the processing of the events to be delayed by several minutes, rendering real-time applications unfeasible. Implementing regularization significantly reduces the number of events, but it also noticeably impacts precision, causing a drop in mAP[0.5].
A mild regularization characterized by a scaling factor of 0.0001 and 0.001 impacts the performance marginally, causing a drop of mAP[0.5] on the chip to 0.868 and 0.780 in the first and second cases, consuming an average power of 33.2 mW and 33.1 mW, respectively. Stronger regularization with a value of 0.005 causes the mAP[0.5] on the chip to drop to 0.565 but also lower significantly the power consumption to 11.8 mW. We determine the best model by giving priority to the highest mAP[0.5], while simultaneously ensuring adherence to the bandwidth limit, which is crucial to prevent any event delay on the chip. Our best model achieves an mAP[0.5] of 0.622 on the chip consuming 19.4 mW.

In Fig.\ \ref{fig:IoU}, the top part illustrates the Intersection over Union (IoU) between correct and predicted bounding boxes for each timestep within a single sample from the on-chip inference, where timesteps are decoded off-chip. 
The middle plot depicts the number of input events as well the number of spikes flowing in the SNN during inference on Speck. When the camera changes direction between saccades, the input events (as shown in the sequence of inputs on the top) and spikes generated decrease accordingly. 
The bottom part shows the power consumption measured on Speck during inference. Overall, we see that the power consumption of Speck during inference is proportional to the number of spikes (i.e., internal activity) flowing in the on-chip SNN, which is itself proportional to the number of input events. If nothing changes in the visual field, the power consumption goes automatically to idle power which is less than 1 mW. This highlights the value of a one-chip neuromorphic solution with an event-based sensor and an asychronous processor, implementing a SNN for low-power always-on inference at the edge.

Finally, in Fig.\ \ref{fig:pareto_synops_plot}, we emphasize the trade-off between the on-chip inference precision and the power consumption. 
It's noteworthy that a higher precision requires a higher number of SynOps/s, resulting in a higher power consumption.

\label{subsec:encoding_decoding}

\section{Discussion and conclusion}
\label{sec:discussion}
In this paper, we presented a first instance of a SNN model deployed on the neuromorphic chip Speck to solve a face detection task. We designed and trained a SNN that is compatible with the hardware constraints of the Speck chip. The model efficiently solves a regression task on-chip in combination with an off-chip decoding linear layer to map output spikes into bounding boxes. We trained the SNN model in simulation using BPTT and a multi-spike activation function, and apllied regularization to control the firing rate of the SNN. After deployment, the best model is selected giving priority to the highest mAP[0.5] while adhering to the bandwidth limitation of the chip cores. This model achieves an on-chip mAP[0.5] of 0.622 while consuming only 19.4 mW.

Our comprehensive investigation yields three key insights that are crucial for the effective deployment of a deep SNN on the Speck processor for solving a complex regression tasks:
\begin{enumerate}
    \item The single-spike activation in simulation is not suitable for on-chip deployment due to significant disparities in the way inputs are processed in simulation versus on the chip. The use of multi-spike activation in simulation for training the model reduces the discrepancy.
    \item To harness multi-spike activation efficiently, a training strategy is introduced involving the normalization of the output spike values before inputting them into a decoding (non-spiking) linear layer. However, layer normalization may introduce limitations and challenges. The firing rate of the spiking layers is not directly controlled by the loss function since all the values are normalized. Moreover, the biases of the non-spiking decoding layer can learn to output stationary bounding box predictions with a sufficiently high IoU with the moving ground truth, even in the absence of output spikes from the model, when the movement in the dataset is minimal. Our study validates that the model does not only rely on the decoding layer biases for predictions.
    \item The understanding of the trade-off between model performance and on-chip power consumption is of crucial importance. We have successfully reduced the model's firing rate by introducing a simple L1 regularization term to the loss function. This penalty on firing rates simultaneously reduces power consumption and model precision, and it prevents the chip stalls due to the high activity exceeding the chip processing limits.
\end{enumerate}

Future research should focus on several key areas. 
First, the training of the model with more complex datasets and utilizing input from the embedded DVS sensors. 
Second, exploring various strategies to penalize the model's firing rate, including setting target firing rates for each layer independently, is of great interest to optimize the trade-off between power consumption and model precision. 
Third, investigating the use of a stateful decoding layer to enhance precision in more complex datasets presents a promising avenue for exploration. 

The question of model latency in relation to synchronous readout remains open. The embedded camera and the per-event processing characteristics of the chip have the potential to achieve very low latency. However, the current system's latency is constrained by the synchronous readout, which still needs to accumulate output spikes. 
The fastest solution would involve increasing the throughput of the readout to match or be similar to the on-chip processing (i.e., per-event prediction) without the necessity of training the model with an extremely small time interval. 
An obvious solution is the use of a FIFO buffer during inference with a length equal to the training window length. This approach allows for decoding the output at a higher frequency by considering values similar to those used during training. 
While this method increases the throughput (i.e., frequency) of bounding box predictions, it does not necessarily guarantee a decrease in detection latency, i.e., the time between pattern appearance and correct detection \cite{Amir_etal17_IBMGesture}. Further investigation is required to explore the relationship between bounding box prediction throughput and actual detection latency.


\section*{Acknowledgment}

The authors would like to express their gratitude to SynSense for their valuable support during the course of this research. SynSense's expertise and resources significantly contributed to the success of our work.


\bibliographystyle{unsrt}
\bibliography{biblio.bib}

\end{document}